\documentclass{article}

\usepackage{arxiv}
\usepackage[utf8]{inputenc} 
\usepackage[T1]{fontenc}    
\usepackage{hyperref}       
\usepackage{url}            
\usepackage{booktabs}       
\usepackage{amsfonts}       
\usepackage{nicefrac}       
\usepackage{microtype}      
\usepackage{lipsum}
\usepackage{times}
\usepackage{url}
\usepackage{latexsym}
\usepackage{multirow}
\usepackage{multicol}
\usepackage{subcaption}
\usepackage{float}
\usepackage{xcolor}
\usepackage{array}
\usepackage{svg}
\usepackage{pifont}
\newcommand{\xmark}{\ding{55}}%
\newcommand{\cmark}{\ding{51}}%
\usepackage{graphicx}
\usepackage[vlined,linesnumbered, plainruled]{algorithm2e}

\newcommand\And{\textbf{and}\xspace}

\SetCommentSty{mycommfont}
\title{Zero-shot hashtag segmentation for multilingual sentiment analysis}

\author{
 Ruan Chaves Rodrigues \\
 Faculty of Informatics \\
  University of the Basque Country\\
  \texttt{ruanchaves93@gmail.com} \\
  \And
 Marcelo Akira Inuzuka \\
 Instituto de Informática \\
  Universidade Federal de Goiás \\
  \texttt{marceloakira@ufg.br} \\
  \And
  Juliana Resplande Sant'Anna Gomes \\
 Instituto de Informática \\
  Universidade Federal de Goiás \\
  \texttt{julianarsg13@gmail.com} \\
  \And
  Acquila Santos Rocha \\
  Instituto de Informática \\
  Universidade Federal de Goiás \\
  \texttt{acquila.santos@gmail.com} \\
  \And
  Iacer~Calixto \\
  Institute for Logic, Language and Computation \\
  University of Amsterdam\\
  \texttt{iacer.calixto@uva.nl} \\
  \And
 Hugo Alexandre Dantas do Nascimento \\
 Instituto de Informática \\
  Universidade Federal de Goiás \\
  \texttt{hadn@inf.ufg.br} \\
}

\begin{document}
\maketitle

\begin{abstract}
Hashtag segmentation, also known as hashtag decomposition, is a common step in preprocessing pipelines for social media datasets. It usually precedes tasks such as sentiment analysis and hate speech detection. For sentiment analysis in medium to low-resourced languages, previous research has demonstrated that a multilingual approach that resorts to machine translation can be competitive or superior to previous approaches to the task. We develop a zero-shot hashtag segmentation framework and demonstrate how it can be used to improve the accuracy of multilingual sentiment analysis pipelines. Our zero-shot framework establishes a new state-of-the-art for hashtag segmentation datasets, surpassing even previous approaches that relied on feature engineering and language models trained on in-domain data.
\end{abstract}


\section{Introduction}
\label{intro}

Word segmentation can be defined as the task of introducing spaces between words when they are not explicitly indicated in the text. One particular case of word segmentation is \textit{hashtag segmentation}, also known as \textit{hashtag decomposition}~\cite{belainine-etal-2016-named}. Hashtags are widely used in social media, whereas hashtag segmentation is often employed as a preprocessing step before applying natural language understanding models for tasks such as sentiment analysis~\cite{boag2015twitterhawk}, hate speech detection~\cite{santosh2019hate} and event detection~\cite{morabia2019sedtwik}. 

Hashtags usually don't bound to standard conventions of written language. They commonly present misspellings, neologisms and previously unknown named entities. Although some hashtags can be easily segmented, a substantial portion of them require models with good generalization performance that can robustly deal with out-of-vocabulary words not seen during training. Some examples that illustrate these problems\footnote{Examples taken from hashtag segmentation dataset Test-BOUN \cite{celebi2016segmenting}.} include: \textit{\#aamirkhan} (`aamir khan'), a Bollywood actor and filmmaker; \textit{\#fangtasyisland} (`fangtasy island'), a misspelling of `fantasy island'; and \textit{\#nooootttttt} (`nooootttttt'), the word `not'.

In this work, we revisit previous hashtag segmentation datasets and show that large-scale general-purpose pretrained language models (LMs) can achieve zero-shot performance equivalent to what has already been obtained by training LMs from scratch for this task.

There are only few studies considering this particular setup, most notably the works of Maddela et al. \cite{maddela2019multi}, Doval et al. \cite{doval2019comparing} and Çelebi et al. \cite{celebi2016segmenting,celebi2018segmenting}. Although previous studies have explored how LMs specifically trained on hashtag datasets can be applied to hashtag segmentation, to the best of our knowledge, none have evaluated the zero-shot performance of pretrained Transformer models on hashtag segmentation tasks.

We developed a simple hashtag segmentation framework that combines two publicly available pretrained LMs, GPT-2 \cite{radford2019language} and BERT \cite{devlin2018bert}, using beam search and re-ranking, and we show that our proposed framework is able to achieve state-of-the-art results on hashtag segmentation datasets.

Our main contributions are:
\begin{itemize}
    \item State-of-the-art results on TEST-BOUN ~\cite{celebi2016segmenting} hashtag segmentation dataset in a zero-shot fashion using publicly available pretrained LMs. To the best of our knowledge, we are the first to investigate how GPT-2~\cite{radford2019language} can be applied to a word segmentation task, and also the first to utilize Transformer models in hashtag segmentation.
    \item A zero-shot approach to hashtag segmentation, that can  seamlessly integrated to natural language processing pipelines in tasks such as multilingual social media sentiment analysis.
    \item An implementation of our framework released as open-source~\footnote{\url{https://github.com/ruanchaves/hashformers}} to reproduce our experiments as well as to implement hashtag segmentation in production environments.
\end{itemize}

The remainder of this paper is organized as follows.
In Section \ref{sec:contextualization}, we discuss the linguistic and conceptual contexts behind the problem of word segmentation, and also relevant related works on hashtag segmentation.
In Section \ref{sec:approach}, we introduce and discuss our approach. In Section \ref{sec:experiments}, we provide details on our experimental setup and present the main experimental findings. Finally, in Section \ref{sec:conclusions}, we draw our conclusions and provide avenues for future work.

\section{Contextualization}
\label{sec:contextualization}

In this section, we discuss the linguistic background behind the task of word segmentation. We also pursue a detailed discussion of how the term \textit{word segmentation} has been utilized in previous research, and we close the section by presenting related work in the field of hashtag segmentation.

\subsection{Linguistic Background}

Not all languages explicitly indicate word boundaries in writing. In European languages, as investigated by Paul Saenger \cite{saenger1997space}, word separation started to be consistently utilized only after the late tenth century. 

Ancient and medieval European manuscripts written before the adoption of word separation are said to be written in \textit{scripta continua}. Thibault Clérice \cite{clerice2020evaluating} has investigated how deep learning architectures can be applied to the word segmentation of \textit{scripta continua} manuscripts in Latin and Old French.

However, word separation has never made its way into several languages outside the Europe. Among these languages, Chinese warrants special mention due to the sheer volume and advanced state of research encountered in the field of Chinese Word Segmentation (CWS). Fu et al. \cite{fu2020rethinkcws} and Li et al. \cite{li2019word} situate CWS in the context of modern deep learning research.

In the last years, the digital revolution has brought about new research fields in word segmentation. Besides hashtag segmentation itself, novel research fields that have arisen from the interaction between human and computers include identifier splitting in source code, as investigated by Rodrigues et al. \cite{rodrigues2020domain} and Razzaq et al. \cite{razzaq2021boostnsift}. 

\subsection{Concepts}
\label{sec:concepts}

Remarkably distinct tasks can be addressed in the literature under the same umbrella term of \textit{word segmentation}. What is meant by word segmentation depends on our definition of what constitutes a \textit{word}, and also on how the writing system of the language we are dealing with indicates word separation, if at all.   

Shao et al. \cite{shao2018universal} following the terminology from the Universal Dependencies framework make a distinction between orthographic words and syntactic words. In his work, \textit{word} is used as a shorthand for \textit{syntactic word}, a syntactic unit that has a unique part-of-speech tag and enters into syntactic relations with other words. From this definition follows \textit{syntactic} word segmentation, the task of identifying the boundaries between spans of part-of-speech tags in a text.

Doval et al. \cite{doval2019comparing}, on the other hand, defines a \textit{word} simply as a sequence of characters delimited by special word boundary characters. A word here is used shorthand for an \textit{orthographic} word. After a text has been corrupted by the removal of its word boundary characters, \textit{orthographic} word segmentation is the task of restoring this corrupted text to a certain orthographic standard by detecting its implicit word boundaries.   

In the context of European languages, syntactic and orthographic word segmentation mean very different tasks. 
\mbox{\textit{`20 000 €'}} is a single syntactic word, since it may receive only one part-of-speech tag (e.g., \textit{CURRENCY}). However, \mbox{\textit{`20 000 €'}} is a sequence of three orthographic words, separated by two word boundary characters.
Conversely, the Spanish \textit{`dámelo'} is a single orthographic word while counting as three syntactic words (\textit{dá}, \textit{me}, and \textit{lo}), each one receiving a distinct part-of-speech tag.

It should be noted that, in the context of CWS and other languages where word separation is completely absent, all word segmentation is inherently syntactic, as word boundary characters are completely absent from orthography and the concept of orthographic word is non-existent.   
In Chinese, word segmentation always follows a certain annotation criterion, which fundamentally depends on how the syntactic analysis of the text is performed. In fact, multi-criteria learning---the technique of developing models capable of adapting to multiple annotation criteria---is an active field of research in CWS, as demonstrated by Huang et al. \cite{huang2020fast} and Ke et al. \cite{ke2020unified}.

In this work, we use the term \textit{word segmentation} as a shorthand for \textit{orthographic word segmentation}, as our goal is to recover word boundaries that have been made implicit in hashtags.
In the experiments presented in this paper, we do not deal with tasks where orthographic word segmentation is not possible, such as in Chinese hashtag segmentation.

\subsection{Related Work}
\label{sec:related}

Madela et al. \cite{maddela2019multi} divide the current approaches for hashtag segmentation in three broad categories: (a) gazetteer and rule-based, (b) word boundary detection and (c) ranking with language model and other features. In the last category, Reuter et al. \cite{reuter2016segmenting} is mentioned as having applied a modified beam search algorithm to English and Brazilian Portuguese hashtag segmentation. Similarly, Doval and Gómez-Rodríguez \cite{doval2019comparing} investigated how a standard beam search algorithm could be applied to word segmentation datasets in multiple European languages.
Çelebi \cite{celebi2016segmenting} initially used gazetteer and rule-based methods for word segmentation, and in subsequent work, Çelebi \cite{celebi2018segmenting} combined feature engineering with language model ranking. 

Although the current state-of-the-art for CWS is mostly dominated by Transformer-based approaches~\cite{huang2019toward}, we could not find any previous research focused on European languages that resorted to these recent architectures.

\section{Approach}
\label{sec:approach}

\begin{figure}[t!]
    \centering
    \includegraphics[width=0.85\textwidth]{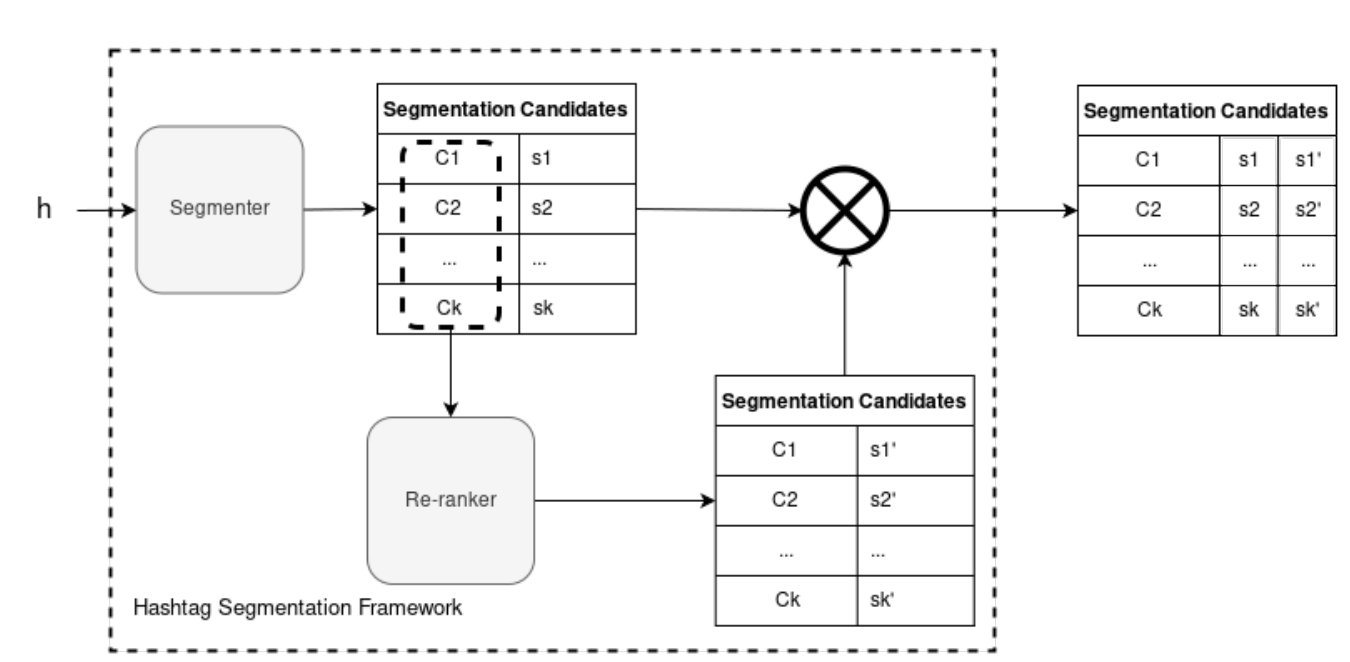}
    \caption{Our proposed framework. Given a hashtag $h$, one \textit{Segmenter} proposes segmentation candidates $C_i$ with scores $s_i$, which are re-ranked by \textit{Re-ranker} that computes new scores $s'_i$ for each $C_i$.}
    \label{fig:framework}
\end{figure}

In this section, our proposed framework is explained in a modular way from the highest to the lowest abstraction level. Initially, we only analyze the output and input of each module, as can be seen in Figure~\ref{fig:framework}. At its highest level of abstraction, the framework receives a hashtag $h$ and produces a list of $k$ target candidates, each one with two scores: $s_i$ and $s_i'$, where $1 \leq i \leq k$. Details of the inputs and outputs of each module are explained below.

\subsection{General Framework}
\label{sec:proposed_framework}

Our framework has two modules: \textit{Segmenter} and \textit{Re-ranker}. The \textit{Segmenter}'s role is to generate a list of target candidates $C_i$ from the hashtag $h$ and produce a score $s_i$ for each one of the candidates. Each candidate $C_i$ has one or more delimiter characters added to $h$. Sequentially in the pipeline, the \textit{Re-ranker} receives this list of candidates and produces a new score $s_i'$ for each one of the candidates. The last operation of the pipeline is a joining of the lists produced by the two previous modules, generating a list of candidates $C_i$ with each one associated with the scores $s_i$ and $s_i'$, coming from the \textit{Segmenter} and \textit{Re-ranker}, respectively.

\subsubsection{Segmenter}
\label{sec:segmenter}

Our approach is primarily based on language modeling and beam search. A language model learns a probability distribution over text sequences $Y = (y_1, \cdots, y_N)$ of any given finite length $N$, where each $y_i$ is a word or token.It is commonly factorized \textit{autoregressively} as below.
\begin{equation}
    P(Y ; \theta) = \prod_{i=1}^{N}{P(y_i | y_{<i}; \theta)}, \label{eq:lm_autoregressive}
\end{equation}
where $\theta$ are the model parameters.
Performing inference with this LM can use a \textit{greedy search} algorithm where one samples the highest probability word $y_i$ at each time given all the previous words $y_{<i}$ insofar.
Beam search is a heuristic search algorithm that is the \textit{de-facto} standard when decoding sequences from models such as the autoregressive LM in Eq.~\ref{eq:lm_autoregressive}. It augments the greedy search algorithm with a \textit{beam} of size $k$. The core idea in beam search is to compute the top-$k$ highest probability words $y_i$ at a time $i$ according to Eq.~\ref{eq:lm_autoregressive} and store each candidate in one of the $k$ beams; for each candidate sequence, one then computes again the highest probability words $y_{i+1}$ for the next time step $i+1$ according to Eq.~\ref{eq:lm_autoregressive}, in other words generating $k^2$ candidates; finally, one keeps only the top-$k$ highest probability candidates, discarding the remaining $k \cdot (k-1)$ candidates.

One of the main advantages of treating word segmentation as language modeling is to make zero-shot transfer possible~\cite{radford2019language}, i.e., to utilize a pretrained LM without the need for re-training or fine-tuning task-specific layers. 

Beam search can be seen as a greedy algorithm that builds a search tree in a breath-first fashion. However, only the best scoring nodes according to a chosen cost function are expanded in each step. Detailed descriptions of how the beam search algorithm can be applied to word segmentation have been presented by Doval and Gómez-Rodríguez \cite{doval2019comparing} and Zhang and Clark \cite{zhang2011syntactic}. We use their general idea of creating a search tree but we use a different expansion method and incorporate a new cost function for pruning tree branches. 

We define data structures $H$, $S$, $T$ and $D$ below to describe our beam search algorithm for hashtag segmentation.
\begin{itemize}
\item $H = <c_1, c_2, ..., c_n >$ is a hashtag with $n$ characters $c_i$, where $n \geq 2$.
\item $S = <c_1,d_1,c_2,d_2,...,c_{n-1},d_{n-1},c_n>$ is a segmentation of H, where $d_i = \epsilon $ represents no word delimiter character and $d_i = \Box$ indicates a delimiter character and $1\leq i \leq n-1$. To refer to a position $j$ in $S$ we use $S_j$, where $1 \leq j \leq (2 \times n)-1$.
\item $T = < S_1, S_2, ..., S_j >$ is a tree of segmentation candidates, where $j$ is the size of the list.
\item $D = \{ <S_1,s_1>, <S_2,s_2>, ...,<S_j,s_j> \}$ is a dictionary of scored segmentation candidates, where $s_i$ is the score of segmentation $S_i$ and $j$ is the size of the dictionary.
\end{itemize}

 We also define five functions that operate on these data structures:
\begin{itemize}
    \item $generate(H)$ : generates $S$ from a hashtag $H$ with all $d_i = \epsilon $.
    \item $length(S)$: returns the size of $S$, including all characters where $d_i = \epsilon$.
    \item $counts(S)$: returns the number of $d_i = \Box$ in S.
    \item $score(T)$: computes $D$ with score $s_i$ for each node $S_i$ in $T$.
    \item $select(D, top_k)$: selects the top-$k$ best scored candidates from $D$. 
    \item $append(T,S)$: include S at the end of T.
\end{itemize}

Finally, we formalize our approach to the beam search algorithm in Algorithm \ref{alg:hsbs}.

\SetStartEndCondition{ }{}{}%
\SetKwProg{Function}{function}{\string:}{}
\SetKwFunction{Range}{range}
\SetKwComment{Comment}{\color{green!50!black}\# }{}
\SetKw{KwTo}{in}\SetKwFor{For}{for}{\string:}{}%
\SetKwIF{If}{ElseIf}{Else}{if}{:}{elif}{else:}{}%
\SetKwFor{While}{while}{:}{fintq}%
\AlgoDontDisplayBlockMarkers\SetAlgoNoEnd\SetAlgoNoLine%
\begin{algorithm}
\caption{Hashtag segmentation beam search ($hsbs$).}\label{alg:hsbs}
\begin{multicols}{3}
\Function{hsbs($H$,$e$,$top_k$)}{
    $D \gets <>$\;
    $S \gets generate(H)$\;
    $T \gets <S>$\;
    \For {$t \in <1,...,e>$}{
        $T = expand(T, t)$\;
        $D = score(T)$\;
        $T = prune(D, top_k)$\;
    }
    $D = score(T)$\;
    \Return{ $D$}
}
\Function{expand($T$, $t$)}{
$T_{expanded} \gets <>$\;
    \For { each  $S \in T$}{
        \If{ $counts(S) >= t -1$ }{
            $l \gets length(S)$\;
            $j \gets 2$\;
            \While{$j < l$}{
                \If{$S_j \neq \Box$}{
                    $S_j \gets \Box $\;
                    append($T_{expanded}, S)$
                }
                $j \gets j + 2$\;
            }
        }
    }
    \Return{ $T_{expanded}$}\;
}
\Function{prune($D, top_k$)}{
    $T_{pruned} \gets <>$\;
    $T_{topk} \gets select(D, top_k)$\;
    \For {$S \in T_{topk}$}{
            $append(T_{pruned},S)$\;
    }
    \Return{ $T_{pruned}$}\;
}
\end{multicols}
\end{algorithm}

\begin{table}[t!]
    \centering
    \begin{tabular}{lc}
    \toprule
    Pseudo-code & Simulation \\
    \midrule

    \begin{minipage}{.3\textwidth}{
    $t = 1$\\
    $T_{expanded} = expand(T)$ 
    }\end{minipage}
    & 
    \begin{minipage}{.5\textwidth}{\includegraphics[width=\textwidth]{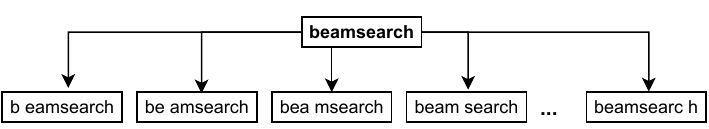}}
    \end{minipage}
    \\
    \midrule
    $D = score(T_{expanded})$ &  \begin{minipage}{.5\textwidth}{\includegraphics[width=\textwidth]{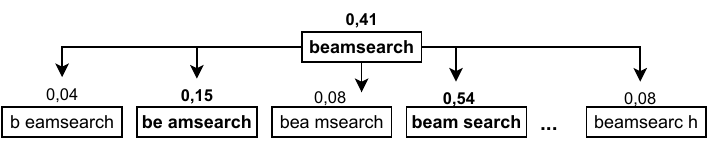}}
    \end{minipage}
    \\
    \midrule
    \begin{minipage}{.3\textwidth}{
    $T_{pruned} = prune(D, 3)$\\
    $t = 2$\\
    $T_{expanded} = expand(T_{pruned})$
    }\end{minipage}
    & \begin{minipage}{.5\textwidth}{\includegraphics[width=\textwidth]{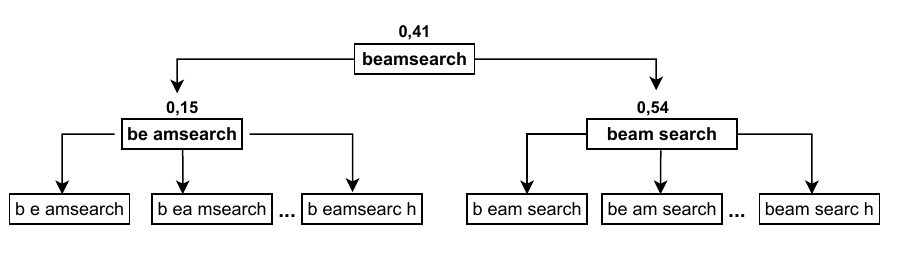}}
    \end{minipage}
    \\
    \bottomrule\\
    \end{tabular}
    \caption{Simulation of the execution of hashtag beam search algorithm.}
    \label{tab:hsbs}
\end{table}

We build a variation of beam search for hashtag segmentation, as presented in Algorithm \ref{alg:hsbs}. In short, the pseudocode consists of: first, a hashtag segmentation tree $T$ is initialized with the hashtag $H$, expanded according to $expand$ function; second, all the nodes in the three are scored by a $score$ function; third, only the $top_k$ selected nodes remain for the next iteration, i.e., the next expansions takes only leaf nodes as inputs.\footnote{Due to the expansion step, generation of descendant nodes, and selection of the fittest individuals, we note the similarity with genetic algorithms. However, since in our algorithm there is no stochasticity nor mutation between individuals (i.e., it is deterministic), it cannot be classified as such.} The number of iterations is limited to $e$ expansions, which correspond to the maximum amount of separator characters to be added to each candidate. An important observation is that the parameters $top_k$ and $e$ must not be excessively small, because if no leaf node is selected for the next iteration, the algorithm may stop abruptly, without proper exploration of the solution space. We determine these values empirically, choosing $\epsilon=13$ and $top_k=20$ for the English language.

Table \ref{tab:hsbs} illustrates the execution of Algorithm
\ref{alg:hsbs} and shows 3 steps: in the first step (iteration $t=1$), the \textit{'beamsearch'} hashtag is expanded into 9 candidates (\textit{'b eamsearch'}, \textit{'be amssearch'}, $\cdots,$ \textit{' beamsearc h'}); in the second step, we generate dictionary $D$ by computing each candidate's score; lastly, the $top_3$ candidates (\textit{'beamsearch'}, \textit{'be amsearch'} and \textit{'beam search'}) are selected with scores 0.41, 0.15 and 0.54, respectively. In iteration $t=2$, leave nodes \textit{'be amsearch'} and \textit{'beam search'} are expanded.

\subsubsection{Re-ranker}
\label{sec:re-ranker}

The Re-ranker receives a list of candidates from the Segmenter, which are the top-$k$ best segmentations selected in the execution of the beam search algorithm. Upon receiving this list, the Re-ranker simply attributes a score to each one of the candidates.
In our implementation, we use BERT \cite{devlin2018bert} as the Re-ranker, and scores are attributed to each candidate via masked language model scoring as defined and implemented by Salazar et al. \cite{salazar2020}.

Our framework's final output is a ranking over candidates selected by the Segmenter and their scores, next to the scores attributed to these candidates by the Re-ranker.

\subsection{Ensembler}
\label{sec:ensembler}

\begin{figure}[t!]
  \centering
  \begin{subfigure}{.45\textwidth}
    \centering
    \includegraphics[width=0.95\textwidth]{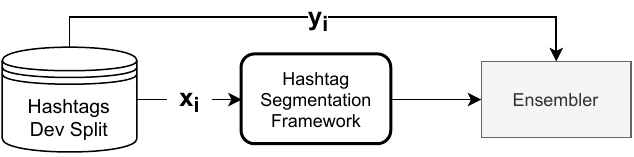}
    \caption{Ensembler training}
    \label{fig:ensembler_training}
  \end{subfigure}
  \begin{subfigure}{.45\textwidth}
    \centering
    \includegraphics[width=0.95\textwidth]{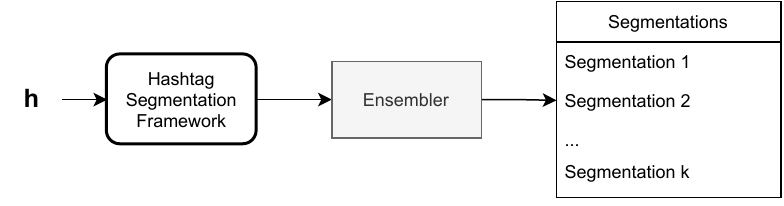}
    \caption{Production pipeline}
    \label{fig:production_pipeline}
  \end{subfigure}
 \caption{\label{fig:esembler_training} In ensembler training (a), $x_i$ and $y_i$ refers to a hashtag and its segmentation from an instance $i$ from Hashtag \textit{Dev} Split. In production pipeline (b), a hashtag $h$ can be segmented in $k$ candidates, where $k \geq 1$}
 \label{fig:ensembler}
\end{figure}

We implement a basic Ensembler as a detachable module applied last in the framework pipeline.
The goal of the Ensembler is to take the ranks provided by the Segmenter and the Re-ranker and combine them into one single best rank. The Ensembler does not integrate the core of our framework as there are several ways to ensemble both ranks, and ideally multiple ensemble options should be taken into account when dealing with a hashtag segmentation problem.   

For the experiments in this paper, we use a simple baseline Ensembler. Given the top two candidates selected by the Segmenter, $c_{1}$ and $c_{2}$, and the scoring functions $f_{S}$ and $f_{R}$ of the Segmenter and the Re-ranker, the decision function $f_{E}$ that characterizes the Ensembler is the defined as below.

$$ f_{E}(c_{1}, c_{2}) = \alpha * |f_{S}(c_{1}) - f_{S}(c_{2})| - \beta * |f_{R}(c_{1}) - f_{R}(c_{2})|$$   

This decision function produces the final rank, and features only two candidates. For any given hashtag and its candidate segmentations, if $f_{E}$ is positive, the Ensembler rank for this hashtag will simply take the top two candidates selected by the Segmenter. If $f_{E}$ is negative, the Ensembler rank will be made of the top two candidates selected by the Segmenter in the order determined by the Re-ranker.

$\alpha$ and $\beta$ are hyperparameters that weight the absolute differences between candidate scores. Both $\alpha$ and $\beta$ are floating point numbers in the range $[0, 1]$ and are optimized through grid search on a development set (illustrated in Figure \ref{fig:ensembler}).
During grid search, we pick values for $\alpha$ and $\beta$ that maximize the F-score achieved on the development set. After tuning, the Ensembler is ready to be integrated into a production pipeline as a module combines the outputs of our hashtag segmentation framework Segmenter and Re-ranker modules.

\section{Experiments}
\label{sec:experiments}

In this section, we describe datasets, our experimental setup and the results obtained by evaluating our hashtag segmentation techniques. In addition to the hashtag segmentation task itself, we perform an extrinsic evaluation by integrating and evaluating our framework into a sentiment analysis pipeline.

\subsection{Datasets}

We concentrate on hashtag datasets that are manually segmented by human annotators and that are associated with state-of-the-art results for a hashtag segmentation task. We therefore use two English hashtag datasets: Test-STAN and Test-BOUN, proposed by Çelebi et al. ~\cite{celebi2016segmenting}. Their respective development sets, Dev-STAN and Dev-BOUN, were utilized for hyperparameter optimization (Section \ref{sec:ensembler}).

\subsection{Experimental setup}

Our experiments were performed on a NVIDIA Tesla V100GPU (with 32 GiBytes of global shared memory). In all our beam search experiments, we considered a fixed beam size $k = 20$ and search tree height $h = 13$.  Candidate scores for GPT-2 were calculated with the \textit{lm-scorer} library~\footnote{\url{https://github.com/simonepri/lm-scorer}} and candidate scores for BERT with the \textit{mlm-scoring}, library~\footnote{\url{https://github.com/awslabs/mlm-scoring}} released by Salazar et al. \cite{salazar2020}. All Transformer models we use are publicly available and have not been further trained or fine-tuned in any way~\footnote{Refer to the HuggingFace Model Hub \url{https://huggingface.co/models}.}.

\subsection{Oracle segmenter evaluation}
\label{subsec:segmenter_evaluation}

\begin{table}[t!]
    \centering
    \resizebox{\textwidth}{!}{%
    \begin{tabular}{rccccccccccccccccc}
    \toprule
        & \multicolumn{8}{c}{\bf Test-STAN} & & \multicolumn{8}{c}{\bf Test-BOUN}\\
        \cmidrule{2-9} \cmidrule{11-18}
        & \multicolumn{2}{c}{Çelebi et al.~\cite{celebi2018segmenting}} & & \multicolumn{2}{c}{GPT-2} & & \multicolumn{2}{c}{BERT} & &  \multicolumn{2}{c}{Çelebi et al.~\cite{celebi2018segmenting}} & & \multicolumn{2}{c}{GPT-2} & & \multicolumn{2}{c}{BERT} \\
        \cmidrule{2-3} \cmidrule{5-6} \cmidrule{8-9} \cmidrule{11-12} \cmidrule{14-15} \cmidrule{17-18}
        & F1 & Acc & & F1 & Acc & & F1 & Acc & & F1 & Acc & & F1 & Acc & & F1 & Acc \\
        N = 1 & \bf 82.9 & \bf 80.4 & & 72.2 & 75.9 & & 43.1 & 41.9 & & \bf 93.2 & \bf 90.0 & & 89.9 & 85.2 & & 62.3 & 57.1 \\
        N = 2 & \bf 92.9 & \bf 91.6 & & 90.7 & 90.2 & & 47.8 & 49.6 & & 96.2 & 94.4 & & \bf 97.9 & \bf 97.0 & & 47.7 & 49.5 \\
        N = 5 & 94.4 & 93.2 & & \bf 97.4 & \bf 97.6 & & 55.4 & 60.6 & & 96.6 & 94.8 & & \bf 99.7 & \bf 99.6 & & 75.0 & 75.6 \\
        N = 10 & 94.4 & 93.2 & & \bf 98.8 & \bf 99.1 & & 62.9 & 69.3 & & 96.6 & 94.8 & & \bf 99.7 & \bf 99.6 & & 79.2 & 80.6 \\
        \bottomrule\\
    \end{tabular}
    }
    \caption{Oracle evaluation on two hashtag segmentation datasets Test-STAN and Test-BOUN where a result is deemed correct in case the gold standard segmentation appears among the N top scored segmentation ($N=\{1,2,5,10\}$). We compare three hashtag segmentation models: the language model proposed in Çelebi et al.~\cite{celebi2018segmenting}, GPT-2 and BERT, the latter two being pretrained LMs used in a strictly zero-shot fashion and without any retraining or tuning. We highlight in bold the best of the three models on each test set. See Section \ref{sec:segmenter} for details on Segmentation algorithms and Section \ref{subsec:segmenter_evaluation} for a detailed discussion of these results.}
    \label{tab:oracle_results}
\end{table}

We compare two publicly available pretrained language models, BERT (\textit{bert-large-uncased-whole-word-masking}) and GPT-2 (\textit{gpt2-large}) and how well they perform when used as Segmenters in a strictly zero-shot fashion without any retraining. We therefore implement our beam search algorithm and test it on the datasets presented by Çelebi et al. \cite{celebi2018segmenting}.

In order to do that, we follow exactly the evaluation procedure proposed by Çelebi et al. \cite{celebi2018segmenting} and consider a result as correct if the gold standard segmentation is among the top scored N segmentations.
We compare the F-score and accuracy for the top N segmentations produced by each model.
As shown in Table \ref{tab:oracle_results}, BERT is ineffective as a Segmenter given that even at the top 10 candidates it does not improve upon previous work \cite{celebi2018segmenting} on Test-STAN and Test-BOUN.    

GPT-2 does not outperform the language model proposed by Çelebi et al. \cite{celebi2018segmenting} when only the top candidate is taken into account, but when using more candidates GPT-2 becomes clearly the best among all models (see Table \ref{tab:oracle_results}). According to the experiments in Maddela et al. \cite{maddela2019multi}, GPT-2's results for the top 10 candidates approaches their human performance. Overall, comparing Çelebi et al. \cite{celebi2018segmenting} and GPT-2, there is an improvement from 94.4\% to 98.8\% in F-score on Test-STAN and from 96.6\% to 99.7\% on Test-BOUN ($N=10$). 

\subsection{Framework evaluation}
\label{sec:framework_evaluation}
    \begin{table}[t!]
        \centering
        \begin{tabular}{llccc}
             \toprule
             \bf Dataset & \bf Architecture & \bf Unsupervised? & \bf F-1 & \bf Accuracy \\
             \midrule
             \multirow{6}{*}{Test-Stanford} & Microsoft Word Breaker \cite{maddela2019multi} & \xmark{} & 84.6 & 83.6 \\
             & Çelebi et al. \cite{celebi2018segmenting} & \xmark{} & 82.9 & 80.4 \\
             & Çelebi et al. \cite{celebi2018segmenting} + feature engineering (FE) & \xmark{} & \textbf{90.2} & 88.5 \\
             & Maddela et al. \cite{maddela2019multi} + feature engineering (FE) & \xmark{} & 89.8 & \textbf{91.0} \\
             & Segmenter (GPT-2) $\rightarrow$ Reranker (BERT), $\alpha=0.0, \beta=1.0$ & \cmark{} & 51.9 & 45.2 \\
             & Segmenter (GPT-2) $\rightarrow$ Reranker (BERT), $\alpha=0.2, \beta=0.1$ & \cmark{} & 85.7 & 84.3 \\
             \midrule
             \multirow{5}{*}{Test-BOUN} & Microsoft Word Breaker \cite{celebi2018segmenting} & \xmark{} & 84.4 & 86.2 \\
             & Çelebi et al. \cite{celebi2018segmenting} & \xmark{} & 93.2 & 90.0 \\
             & Çelebi et al. \cite{celebi2018segmenting} + feature engineering (FE) & \xmark{} & 94.9 & 92.9 \\
             & Segmenter (GPT-2) $\rightarrow$ Reranker (BERT), $\alpha=0.0, \beta=1.0$ & \cmark{} & 72.7 & 62.3 \\
             & Segmenter (GPT-2) $\rightarrow$ Reranker (BERT), $\alpha=0.2, \beta=0.1$ & \cmark{} & \textbf{95.6} & \textbf{93.4} \\
             \bottomrule \\
        \end{tabular}
        \caption{F-score and accuracy achieved by our framework on the Test-Stanford~\cite{celebi2016segmenting} and Test-BOUN~\cite{celebi2016segmenting} datasets. We compare to previous results reported in the literature, and investigate searching for best hyperparameters on the dev-set ($\alpha=0.2, \beta=0.1$) and blindly using BERT for re-ranking ($\alpha=0.0, \beta=1.0$). More details are available in Section~\ref{sec:framework_evaluation}. We surpass the Microsoft Word Breaker baseline on Test-Stanford and achieve new state-of-the-art results on the Test-BOUN dataset.
        }
        \label{tab:reranking_results}
    \end{table}

Now that we demonstrated the effectiveness of GPT-2 as a Segmenter in Section \ref{subsec:segmenter_evaluation}, we concentrate our efforts on re-ranking its candidates. Table \ref{tab:reranking_results} describes our experiments with two distinct ways of re-ranking the top 2 candidates selected by GPT-2. In both methods, the algorithm decides between the original rank presented by GPT-2 and the re-ranking proposed by BERT according to the candidate scores and the weights $\alpha$ and $\beta$.

Our first method is a baseline where we decide to always trust the re-ranking proposed by BERT. In our Ensembler module, this can effectively be achieved by setting $\alpha$ to zero and $\beta$ to one or any other arbitrary value above zero.
In our second method we determine $\alpha$ and $\beta$ by grid search on the development set for each one of the datasets, i.e., Dev-Stanford and Dev-BOUN. 
Results in Table \ref{tab:reranking_results} indicate that blindly trusting BERT for re-ranking is outperformed even by the standard Microsoft Word Breaker baseline. However, if the weights $\alpha$ and $\beta$ are determined according to the development set, our framework improves upon the Word Breaker baseline on Test-Stanford, and is outperformed only by solutions that rely on feature engineering.
Finally, we note that our framework achieves a new state-of-the-art for the Test-BOUN dataset without any feature engineering (Table \ref{tab:reranking_results}), outperforming the previous combination of a language model and feature engineering (proposed in Çelebi et al. \cite{celebi2018segmenting}).

\subsection{Extrinsic evaluation}

\begin{table}[t!]
    \centering
    \scalebox{0.9}{
    \begin{tabular}{r|ccc|ccc|ccc}
        \toprule
        & \multicolumn{3}{c}{\textbf{T}} & \multicolumn{3}{c}{\textbf{CMT}} & \multicolumn{3}{c}{\textbf{CMTS}} \\
        & Acc & Recall & F1 & Acc & Recall & F1 & Acc & Recall & F1 \\
        \midrule
        \multicolumn{10}{c}{Arabic $\rightarrow$ English}\\
        \midrule
        DistilBERT-SST2 &           67.6 &           67.6 &           67.5 &  67.1 &    67.1 &  66.7 &  69.0 &  69.0 &  68.8 \\
        RoBERTa-SST2 &           72.9 &           72.9 &           72.9 &  73.1 &    73.1 &  73.1 &  73.6 &  73.6 &  73.6 \\
        RoBERTa-multiple &           \underline{73.4} &          \underline{73.4} &           \underline{73.1} &  \underline{76.6} &    \underline{76.6} & \underline{76.2} &  {\bf 76.7} &  {\bf 76.7} &  {\bf 76.4} \\
        \midrule
        \multicolumn{10}{c}{German $\rightarrow$ English}\\
        \midrule
        DistilBERT-SST2 &  69.8 &    69.8 &  68.7 &           70.9 &           70.9 &           69.8 &  71.2 &  71.2 &  70.2 \\
        RoBERTa-SST2 &  79.3 &    79.3 &  79.2 &           80.3 &           80.3 &           80.3 & 80.9 &  80.9 & 80.8 \\
        RoBERTa-multiple &  \underline{83.1} &    \underline{83.1} &  \underline{83.1} &  \textbf{84.1} &  \textbf{84.1} &  \textbf{84.1} &           \underline{83.6} &           \underline{83.6} &           \underline{83.6} \\
        \midrule
        \multicolumn{10}{c}{ Hindi $\rightarrow$ English}\\
        \midrule
        DistilBERT-SST2 &           52.4 &           52.4 &           52.3 &           53.4 &           53.4 &           53.4 &  53.4 &  53.4 &  53.4 \\
        RoBERTa-SST2 &  \textbf{56.0} &  \textbf{56.0} &  \textbf{56.0} &           \underline{54.3} &           \underline{54.3} &           \underline{54.3} &           \underline{54.1} &           \underline{54.1} &           \underline{54.1} \\
        RoBERTa-multiple &           52.6 &           52.6 &           52.5 &           54.0 &           54.0 &  53.9 &  54.0 & 54.0 &           53.8 \\
        \midrule
        \multicolumn{10}{c}{ Italian $\rightarrow$ English }\\
        \midrule
        DistilBERT-SST2 &           73.8 &           73.8 &           73.8 &           72.6 &           72.6 &           72.4 &  77.4 &  77.4 &  77.4 \\
        RoBERTa-SST2 &           \underline{78.1} &           \underline{78.1} &           \underline{77.9} &  \textbf{81.0} &  \textbf{81.0} &  \textbf{80.9} &           \underline{80.7} &           \underline{80.7} &           \underline{80.5} \\
        RoBERTa-multiple &           77.6 &           77.6 &           77.2 &  79.7 &  79.7 & 79.3 &           78.1 &           78.1 &           77.6 \\
        \midrule
        \multicolumn{10}{c}{ Spanish $\rightarrow$ English}\\
        \midrule
        DistilBERT-SST2 &           78.3 &           78.3 &           78.3 &           79.1 &           79.1 &           79.1 &  79.7 &  79.7 &  79.7 \\
        RoBERTa-SST2 &           81.0 &           81.0 &           81.0 &           82.1 &           82.1 &           82.1 &  82.6 &  82.6 &  82.6 \\
        RoBERTa-multiple &  \textbf{85.0} &  \textbf{85.0} &  \textbf{85.0} &           \underline{84.8} &           \underline{84.8} &           \underline{84.8} &           \underline{84.0} &           \underline{84.0} &           \underline{83.9} \\
        \bottomrule\\
    \end{tabular}
    }
    \caption{Results for English sentiment analysis models applied on tweets taken from the datasets on the Unified Multilingual Sentiment Analysis Benchmark \cite{barbieri2021xlmtwitter}. All datasets were automatically translated into English before applying the models.
    {\bf DistilBERT-SST2} \cite{sanh2019distilbert} uses knowledge distillation and is fine-tuned on the SST-2 sentiment analysis dataset (\textit{distilbert-base-uncased-finetuned-sst-2-english}).
    {\bf RoBERTa-multiple} \cite{heitmann2020} is a RoBERTa model fine-tuned on multiple sentiment analysis datasets, including tweets (\textit{siebert/sentiment-roberta-large-english}).
    {\bf RoBERTa-SST2} \cite{morris2020textattack} is a RoBERTa model fine-tuned using adversarial training (\textit{textattack/roberta-base-SST-2}).
    We {\bf bold-face} the best results per language group, and \underline{underscore} the best result per experimental setup ({\bf T}, {\bf CMT}, {\bf CMTS}) in each language group. {\bf T (Translation)}: We simply translate the tweet. {\bf CMT (Code-Mixed Translation)}: We translate the tweet and the segmented hashtag separately into English. {\bf CMTS (Code-Mixed Translation and Segmentation)}:  After performing {\bf CMT}, we restore the spaces between the words on the hashtag.}
    \label{tab:all_langs_sentiment_analysis}
\end{table}

We now apply our  hashtag segmentation framework on the task of Twitter sentiment analysis, so as to demonstrate its application in a practical setting. In these experiments, we consider the subset of positive and negative tweets in the Unified Multilingual Sentiment Analysis Benchmark (UMSAB) \cite{barbieri2021xlmtwitter}, while ignoring tweets labelled as neutral. We make this decision so that we can perform our experiments in an entirely zero-shot setting with a larger variety of publicly available language models.

\begin{table}[t!]
    \centering
    \resizebox{\textwidth}{!}{
    \begin{tabular}{lllcc}
         \toprule
         \bf Language & \bf Segmenter & \bf Reranker & \bf spaCy Tokenizer & \bf Translator \\
         \midrule
         Arabic & aubmindlab/aragpt2-large & aubmindlab/bert-large-arabertv2 & xx\_sent\_ud\_sm & opus-mt-ar-en \\
         German & dbmdz/german-gpt2 & bert-base-german-cased & de\_dep\_news\_trf & opus-mt-de-en \\
         Hindi  & surajp/gpt2-hindi & ai4bharat/indic-bert & xx\_sent\_ud\_sm & opus-mt-hi-en \\
         Italian & GroNLP/gpt2-small-italian & dbmdz/bert-base-italian-xxl-cased & it\_core\_news\_lg & opus-mt-it-en \\
         Spanish & mrm8488/spanish-gpt2 & dccuchile/bert-base-spanish-wwm-cased & es\_dep\_news\_trf & opus-mt-es-en \\
         \bottomrule \\
    \end{tabular}
    }
    \caption{Models used in extrinsic evaluation experiments. Segmenter, Re-ranker and Translator are referred by their identifier codes on the HuggingFace Hub, and spaCy tokenizers by their names under the spaCy library version 3.0.}
    \label{tab:model_list_translation}
\end{table}

For each one of the languages listed on Table \ref{tab:model_list_translation}, we choose the largest GPT-2 and BERT models made publicly available at the time of this publication, which are used respectively as Segmenter and Re-ranker. 
For these particular experiments, we do not adjust $\alpha$ and $\beta$ on each development set. Rather, we use $\alpha = 0.2$ and $\beta = 0.1$, which have empirically been demonstrated to achieve acceptable results on the intrinsic evaluation of our hashtag segmentation system, as shown in Table \ref{tab:reranking_results}.

Next, for each language in the UMSAB benchmark in Table \ref{tab:all_langs_sentiment_analysis} we segment all the hashtags on the test set using the appropriate language models listed in Table \ref{tab:model_list_translation} as Segmenter and Re-ranker.
After obtaining segmentation outputs for all the hashtags, we automatically translate the datasets into English through three distinct methods: 

\begin{itemize}
    \item \textbf{Translation (T)}: We simply submit the original tweets to MarianMT~\footnote{https://huggingface.co/docs/transformers/model\_doc/marian} automatic translation engine, without resorting to hashtag segmentation during any step of the process. This method is illustrated in Figure.
    \begin{figure}[t!]
    \centering
    \includegraphics[width=0.9\textwidth]{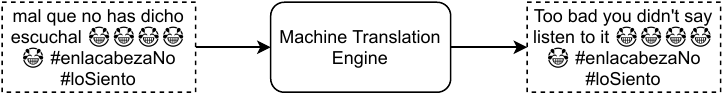}
    \caption{Translation (T). A Spanish tweet is translated into English with a Machine Translation Engine.}
    \label{fig:translation_method_t}
    \end{figure}
    \item \textbf{Code-Mixed Translation (CMT)}: All hashtags in the original tweet are segmented, translated and rejoined to produce a translated hashtag, effectively resulting in a code-mixed tweet where the hashtags are in English, our target language, and the rest of the tweet is in the source language. This code-mixed content is then submitted to an automatic translation engine. This method is illustrated in Figure \ref{fig:translation_method_cmt}.
    \begin{figure}[t!]
    \centering
    \includegraphics[width=0.9\textwidth]{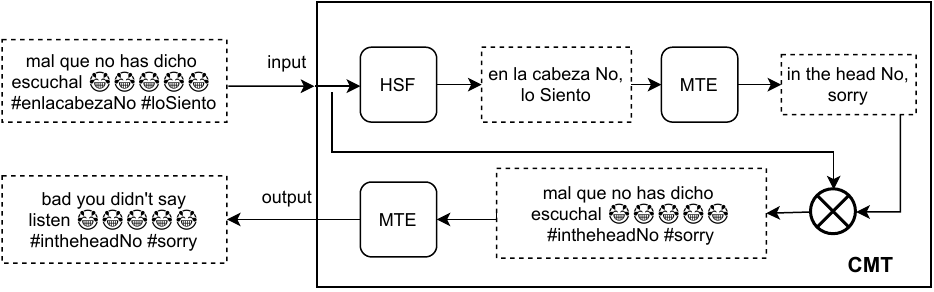}
    \caption{Code-Mixed Translation (CMT). We use our hashtag segmentation framework (HSF) to segment hashtags. The machine translation engine (MTE) is used twice separately, to translate the tweet and the segmented hashtags.}
    \label{fig:translation_method_cmt}
    \end{figure}

    \item \textbf{Code-Mixed Translation and Segmentation (CMTS)}: This method introduces an additional step to CMT. After retrieving the translated output from the CMT method, we recover the spaces that were present before the words on the hashtags were rejoined. This method is illustrated in Figure \ref{fig:translation_method_cmts}.
    \begin{figure}[t!]
    \centering
    \includegraphics[width=0.9\textwidth]{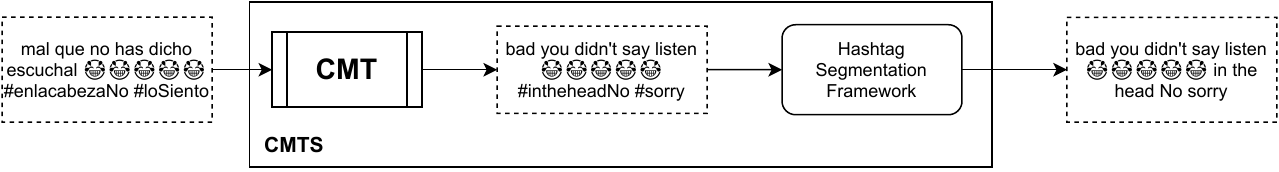}
    \caption{Code-Mixed Translation and Segmentation (CMTS). Here we add an additional step to Code-Mixed Translation (CMT), illustrated in Figure \ref{fig:translation_method_cmt}.}
    \label{fig:translation_method_cmts}
    \end{figure}

\end{itemize}

The Translation method serves as a baseline against which we should compare the CMT and CMTS methods, which make use of our hashtag segmentation system.  
As Table \ref{tab:all_langs_sentiment_analysis} indicates, we could not find a method that consistently ranked above simply translating the tweets for all possible languages.
In instances where our hashtag segmentation solution was effective, significant gains were achieved over the translation baseline. In Arabic, the CMTS method achieved a best F-score of 76.4\%, which is 3.3\% above the baseline. Similar performance achievements occurred in languages which are more closely related to English, such as German and Italian.
For the Spanish dataset, considering the best performing model RoBERTa-multiple, the CMT and CMTS methods scored respectively 0.2\% and 1.0\% below the translation baseline. The Hindi dataset was the most challenging for the sentiment analysis models, and the results similarly could not be improved by our translation methods.
Such variations in performance can be attributed to the distinct pre-training procedures of the models listed in Table \ref{tab:model_list_translation} and the particular characteristics of each dataset. 


\section{Conclusions and future work}
\label{sec:conclusions}

We propose a zero-shot framework that uses pretrained Transformer architectures for hashtag segmentation, and show that it is either competitive or superior to previous approaches based on language models trained from scratch and feature engineering. 

Our zero-shot framework allows us to take a multilingual approach to sentiment analysis without the additional computational costs of training language models for hashtag segmentation in each one of our source languages. 

In order to explore the full potential of our framework, further experiments should consider improvements such as deploying multiple Re-ranker modules and using more sophisticated ensembling techniques, especially those that are capable of taking into account all candidates proposed by the Segmenter module.

Furthermore, we believe conclusive experiments should be made by systematically trying improve the state-of-the-art for each one of the datasets on the Unified Multilingual Sentiment Analysis Benchmark (UMSAB) through the addition of hashtag segmentation to multilingual sentiment analysis pipelines.

We hope this paper will foster future discussions on the applications of hashtag segmentation in social media sentiment analysis, which has been relatively unexplored in comparison with more active fields of research in word segmentation.

\section*{Acknowledgements}

We thank Deep Learning Brasil and LaMCAD/UFG for providing the computer resources for this research.
IC has received funding from the European Union’s Horizon 2020 research and innovation programme under the Marie Skłodowska-Curie grant agreement No 838188.

\bibliographystyle{unsrt}
\bibliography{references}

\end{document}